\documentclass[final]{cvpr}

\usepackage{times}
\usepackage{epsfig}
\usepackage{graphicx}
\usepackage{amssymb}
\usepackage{amsmath}
\usepackage{cuted}
\usepackage{capt-of}
\usepackage{soul}
\usepackage{wrapfig}
\usepackage{overpic}

\usepackage[pagebackref=true,breaklinks=true,colorlinks,bookmarks=false]{hyperref}

\def\cvprPaperID{8562} 
\def\confYear{CVPR 2021}

\newcommand{\seclabel}[1]{\label{sec:#1}}

\def\eg{\emph{e.g.}} 

\def\ie{\emph{i.e.}}

\def\etal{\emph{et al.}}

\newcommand{\norm}[1]{\left\lVert#1\right\rVert}

\def\point3{\bar{X}}

\def\poincare/{Poincar\'e}

\def\eg{\textit{e.g.}}
\def\etal{\textit{et al.}~}

\def\ie{\textit{i.e.}}

\newcommand{\abs}[1]{\left\vert#1\right\vert}

\newcommand{\parr}[1]{\left (#1\right )}

\newcommand{\Real}{\mathbb R}

\def\mobius/{M{\"o}bius}

\newcommand{\beginsupplement}{%
        \setcounter{section}{0}
        \renewcommand{\thesection}{A}%
        \setcounter{subsection}{0}
        \renewcommand{\thesubsection}{A.\arabic{subsection}}
        \setcounter{table}{0}
        \renewcommand{\thetable}{A\arabic{table}}%
        \setcounter{figure}{0}
        \renewcommand{\thefigure}{A\arabic{figure}}%
     }

\begin{document}

\title{Neural Surface Maps
    }

\author{
Luca Morreale$^{1}$ 
\and
Noam Aigerman$^{2}$
\and
Vladimir Kim$^{2}$
\and
Niloy J. Mitra$^{1,2}$\\
\and
$^{1}$University College London \quad
\and
$^{2}$Adobe Research
}

\maketitle

\begin{abstract}
Maps are arguably one of the most fundamental concepts used to define and operate on manifold surfaces in differentiable geometry. Accordingly, in geometry processing, 
maps are ubiquitous and are used in many core applications, such as paramterization, shape analysis, remeshing, and deformation. Unfortunately, most computational representations of surface maps do not lend themselves to manipulation and optimization, usually entailing hard, discrete problems. While algorithms exist to solve these problems, they are problem-specific, and a general framework for surface maps is still in need. 

In this paper, we advocate considering neural networks as encoding surface maps. Since neural networks can be composed on one another and are differentiable, we show it is easy to use them to define surfaces via atlases, compose them for surface-to-surface mappings, and optimize differentiable objectives relating to them, such as any notion of distortion, in a trivial manner. 
%
%
In our experiments, we represent surfaces by generating a neural map that approximates a UV parameterization of a 3D model. Then, we compose this map with other neural maps which we optimize with respect to distortion measures. We show that our formulation enables trivial optimization of rather elusive mapping tasks, such as maps between a collection of surfaces. 
\end{abstract}

\section{Introduction}

\begin{figure}
\begin{overpic}[width=\columnwidth]{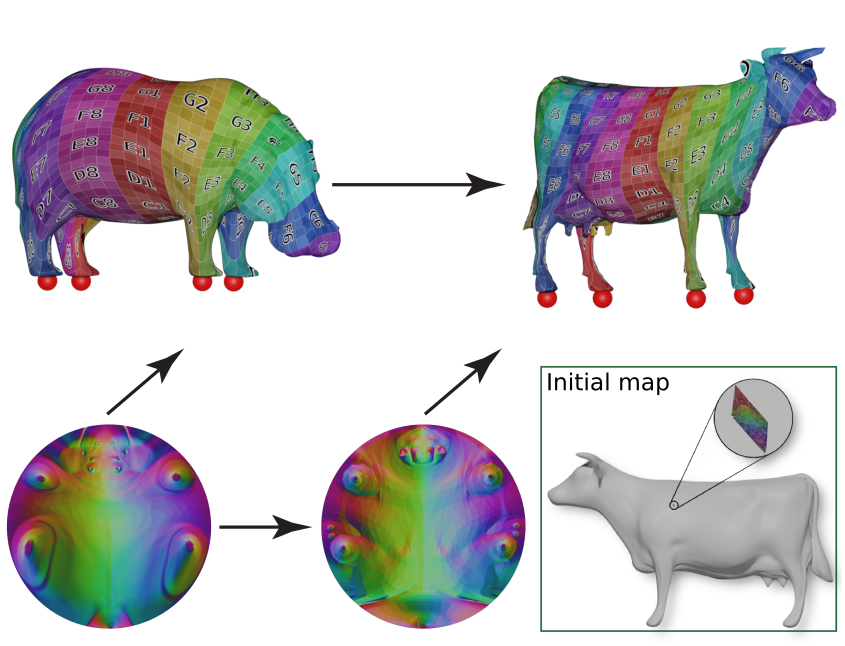}
\huge
\put(46,57){$f$}
\put(9,31){$\phi$}
\put(46,31){$\psi$}
\put(28,18){$h$}
\end{overpic}
    \caption{Two surfaces are respectively represented by two neural maps, $\phi$ and $\psi$, each mapping the unit square to 3D. A distortion-minimizing surface-to-surface map $f$ is visualized by texture transfer. This surface map is achieved by a third neural map $h$ between the square to itself, which yields the surface map by an implicit composition of the three neural maps. The distortion of the composed map $f$ is trivially optimized by defining it as a loss in pytorch and optimizing with respect to $h$. Inset shows the initial random map of the hippo to the cow is a poor initialization, covering only a very tiny region of the cow model. Optimization results in a final map that covers the whole target surface, and reducing isometric distortion, resulting in an average Symmetric Dirichlet energy of $11$ between these highly-nonisometric surfaces. As input, 4 keypoint constraints were used, each on one leg.}
    \label{fig:teaser}
 \end{figure}
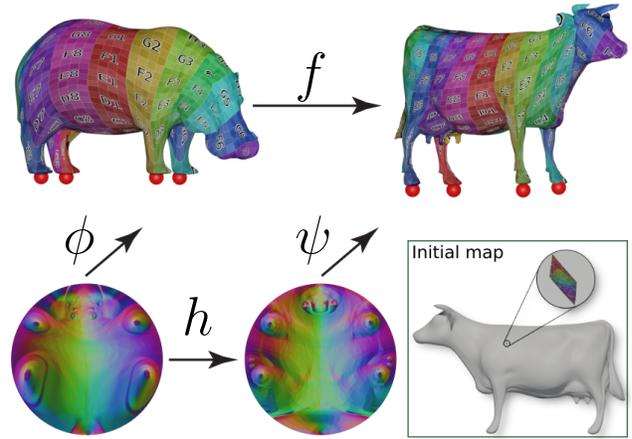

Maps are one of the most fundamental concepts in surface geometry: in differential geometry, a surface, \ie, a 2-manifold, is usually (locally) defined as the image of a (non-degenerate) map
$$
f:\Real^2\to\Real^n.
$$
Not surprisingly, maps are also used to define correspondences between different parts of surfaces in an atlas, to evaluate similarity between surface pairs, or across surface collections. 

Accordingly, computing maps is central in most geometry processing tasks operating on surfaces. The ubiquitous concept of a UV map \cite{sheffer_mesh_2006}, mapping a surface into the plane,  provides a local coordinate system on surfaces, and hence enables downstream tasks such as texturing, surface correspondence, remeshing, quad-meshing~\cite{quadSurvey:2013} to name only a few. Similarly, surface-to-surface maps \cite{Schreiner:2004:IM:1015706.1015812} enable defining correspondences between surfaces, which are at the heart of shape analysis, transfer of properties, deformations, or defining morph sequences. Indeed, almost all shape processing tasks, including parameterization, surface correspondence,  remeshing, and deep learning on surfaces, heavily rely  on access to such surface maps. 

However, many of the tasks related to maps and their computation become extremely hard  to handle when the target domain is a surface, \ie, a 3D mesh ($n=3$). This is due mostly to the fact that meshes are combinatorial representations, which in turn leads to a combinatorial representation of the surface maps, and taints the optimization task with a combinatorial nature as well. Although elegant solutions in the form of discrete differential geometry~\cite{praun2001consistent,schmidt2020inter}, meshing invariant spectral analysis~\cite{melzi_zoomout_2019,mandad2017variance}, functional maps~\cite{ovsjanikov_functional_2012,litany_deep_2017} have been proposed to work around the combinatorial representation, the diverse choices and different data representations inhibit easy end-to-end optimization and adaptation outside the specialized geometry processing community. 

As an example,  consider the problem of computing a mesh-to-mesh mapping in which a continuous map from one surface to the other is computed: one needs to account for the image of each source vertex, which lands on a triangle of the other mesh, and the image of a source edge may span several triangles of the target; this leads to extensive bookkeeping, and any attempt to optimize, \eg, the map's inter-surface distortion leads to combinatorial optimization of the choice of target triangle for each source vertex as in \cite{Schreiner:2004:IM:1015706.1015812,Kraevoy:2004:CCR:1015706.1015811}. An alternative is to optimize proxy maps into a common base domain \cite{aigerman_lifted_2014} in the hope that the resulting surface-to-surface map will be optimized by proxy. Such an approach, however, does not yield surface maps that are even a local minimizer of the energy they set to minimize. This is particularly problematic when optimizing inter-surface maps across shape collections.

In this work, we consider \emph{neural networks} as a parametric representation of both individual surfaces as well as inter-surface maps. Specifically, we consider  networks with parameters $\theta$ that receive 2D points as input and output points either in 2D or 3D, $\phi_\theta:\Real^2\to\Real^n$. While this definition is similar to, \eg, AtlasNet~\cite{groueix_papier-mache_2018}, we do not aim to perform any learning task, and our network does nothing more than map 2D points with the aim of performing one task: approximate a single surface map $\phi_\theta
\sim f:\Real^2\to\Real^n$, so we can work with neural networks instead of with, e.g., mappings of triangular meshes. 

Specifically, we use a map $\phi_\theta \sim f:\Real^2\to\Real^3$ to directly characterize a given manifold shape (restricted to surface patches homeomorphic to a disc), and use another map $\psi_\beta \sim g:\Real^2\to\Real^2$ to update the surface map by restricting movements on the underlying 2-manifold. 
These neural networks are, by construction, differentiable and composable with one another, hence they lend us a simple model for defining a differentiable algebra of surface maps, enabling us to compose maps with one another and optimize objectives directly over their composition, rather than propose approximations via intermediate proxy domains.

We employ this concept in two ways that build on top of one another: first, we revisit the differential-geometry definition of a surface as  a map from 2D to 3D, by overfitting a neural network to a given UV parameterization computed via a standard parameterization algorithm, such as Tutte's embedding \cite{tutte1963draw} or SLIM \cite{rabinovich2017scalable}. Two such maps, $\phi,\psi$, are shown in Figure \ref{fig:teaser}. This gives us a parametric, differentiable representation of the surface, from a canonical domain. Second, we compose the overfitted map with other maps, either to optimize the distortion of the map, or to compute distortion-minimizing maps between two or more surfaces. Figure~\ref{fig:teaser} shows an example of a distortion-minimizing map $f$ defined by composing $h$ with $\phi,\psi$.

We evaluate our method on a variety of triangular meshes with varying complexity and show their efficacy in computation of parameterizations, surface-to-surface distortion-minimizing mapping, and also for mapping across collections of shapes. We also provide comparison to baseline methods. In summary, our main contribution is introducing neural surface map as a  novel representation and utilizing it towards addressing a variety of geometry processing applications. We particularly stress the modular nature of the representation that enables harnessing the power of current deep learning frameworks to solve many (classical) shape analysis tasks in a uniform framework. Code available from the project page \url{http://geometry.cs.ucl.ac.uk/projects/2021/neuralmaps/}.

\section{Related Works} \seclabel{literature}
\begin{figure*}[t!]
    \centering
    \begin{tabular}{cc}
        \includegraphics[width=\textwidth]{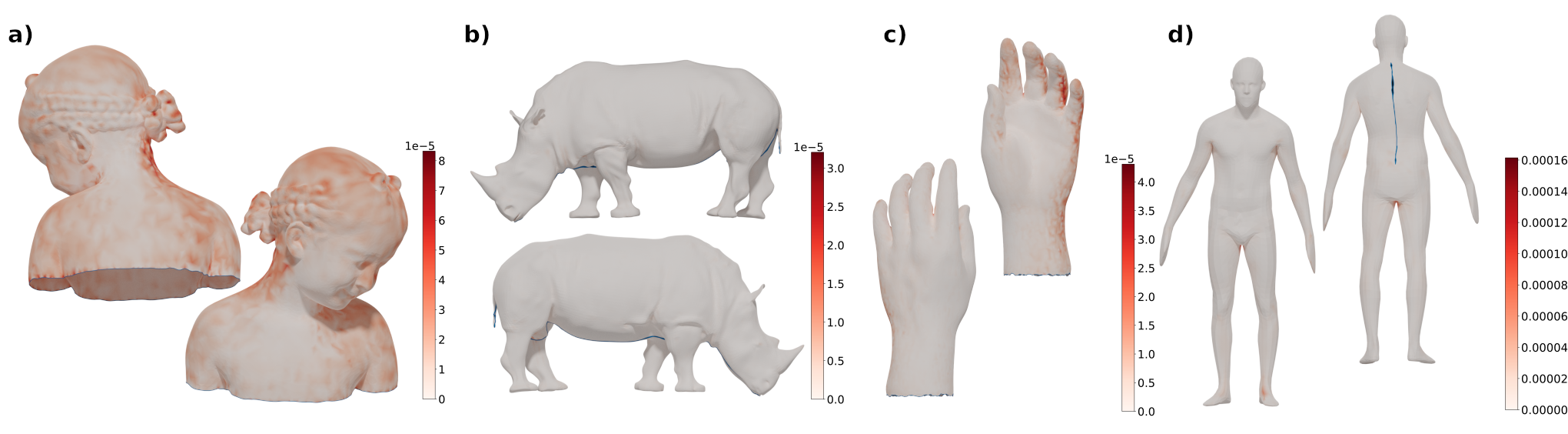} 
    \end{tabular}
    \caption{Parameterizations of the Bimba~(a), rhino~(b), Tosca hand~(c) and FAUST~(d) models represented via overfitted neural maps. The models are colored based on deviation from source models. Distortions, if any, introduced by the mappings are not considered at this stage. }
    \label{fig:bimba_nueral_map}
\end{figure*}
\subsection{Surface maps}

Mappings of surfaces (mostly, meshes) is an active research area.  Usually, algorithms compute surface maps by striving for a specific type of map, such as a harmonic one \cite{tutte1963draw,Pinkall93computingdiscrete,floater2003one,gortler2006discrete,aigerman_orbifold_2015}. In other cases the goal is to compute a map that minimizes or bounds some notion of distortion such as conformality - preservation of angles - \cite{levy2002least,kazhdan2012can,Lipman:2012:BDM}, or isometry - preservation of local distances - \cite{sorkine2007rigid,aigerman_lifted_2014,rabinovich2017scalable,weber2014locally}.  
Many surface mapping algorithms focus specifically on parameterizations, see  \cite{floater_surface_2005,sheffer_mesh_2006} for a more detailed review. On the other hand, the task of computing mappings between two surfaces is a long-studied and notoriously hard problem. Our work uses the popular concept of a common base domain to which two surfaces are mapped, to define the surface map via the overlay of the two maps in the base domain, which can either be a coarse mesh \cite{Lee:1999,Schreiner:2004:IM:1015706.1015812,Kraevoy:2004:CCR:1015706.1015811,Bradley08} or a planar domain \cite{aigerman2014,weber2014locally,aigerman2015,aigerman_hyperbolic_2016}. The resulting surface map can be optimized via direct mesh-mesh intersection at the price of yielding an optimization problem with a significant combinatorial part \cite{Schreiner:2004:IM:1015706.1015812}, due to the discreteness of the triangles. Otherwise, the properties of the resulting map are ignored at the hope that optimizing the maps into the common domain will be sufficient \cite{aigerman_lifted_2014}. 

Soft notions of maps such as Functional Maps \cite{ovsjanikov2012functional} enable a parametric definition of fuzzy maps which can be used in a deep learning context \cite{litany_deep_2017,  donati2020deep}, however they define fuzzy correspondences and not a continuous surface map from one surface to another.

Cycle-consistency \cite{nguyen2011optimization,huang2013consistent} is a trait of maps between a collection of surfaces, that ensures that the set of surfaces are in global correspondence by ensuring that mapping a point across random shapes until reaching the source shape maps the point back to itself. We present the first method to minimize the distortion of a cycle-consistent collection of maps.

\subsection{Neural surface representation}

Atlas-based representations have been prevalent in geometrical deep learning, mostly focusing on generative and analysis tasks (e.g. segmentation), but not on raw representation of surface mapping. 

Many works consider UV-mapping a 3D surface, rendering surface functions as a 2D image, and applying a neural network to the images \cite{sinha2016deep,Maron17,benhamu2018multichart,haim2019surface}. Representing surfaces as regularly-sampled image grids suffers from high distortions in the mapping, where large surface areas can be mapped to sub-pixel regions. Some works  apply neural networks directly on meshes \cite{hanocka_meshcnn_2018,liu2020neural}. Generative techniques, such as FoldingNet~\cite{yang_foldingnet_2018} and AtlasNet~\cite{groueix_papier-mache_2018} propose to alleviate this issue by using a neural network to approximate the atlas map from 2D to 3D as a continuous function conditioned on a latent shape code. Many extensions have been proposed including regularizing differential properties of the mapping~\cite{bednarik_shape_2019}, optimizing for elementary shape of the atlas~\cite{deprelle_learning_2019}, and forcing surfaces to align with the level set of shape's implicit function~\cite{Poursaeed20a}.

While our network architectures are inspired by these prior techniques, as we also train a neural module that maps 2D points to mesh surfaces, they have very different underlying objectives. The aforementioned methods aim at training a network to reconstruct shapes, conditioned on a latent code. Our work is distinct from them, in that it focuses on considering each neural network as a unique, single surface map, and using this representation to solve classic geometry processing problems in surface mapping.

\subsection{Neural shape representation}

In addition to surface-based representations many alternatives have been used, such as voxels~\cite{liu2018voxelgan,DBLP:journals/corr/GirdharFRG16,DBLP:journals/corr/BrockLRW16,dai2017complete}, point clouds~\cite{achlioptas2018latent_pc,Su2017PointGen}, meshes~\cite{dai2019scan2mesh}, and implicit functions~\cite{park_deepsdf_2019}. From these techniques only the neural implicit representations do not suffer from discretization artifacts, since they use neural modules to represent continuous functions, mapping a point in 3D to an occupancy value. Similarly to surface-based methods they aim to create a shared latent space for all shapes, and various extensions have been proposed, such as enforcing unit gradient to satisfy the Eikonal equation~\cite{gropp_implicit_2020}, and sign-agnostic version of this normalization~\cite{atzmon_sal_2020,atzmon_sal_2019}. Littwin and Wolf in \cite{littwin_deep_2019} introduce a meta-learning approach closely related to HyperNetworks~\cite{ha_hypernetworks_2016} for implicit representations. A meta-model $f$ regress weights $\theta_g$ for the implicit function $g$, which describes the signed distance field for a specific shape. Although the approaches introduced above are very successful, they focus on generalizing over a plethora of models rather than represent a single one. Hence, they all present artefacts and thus can only be used as a rough proxy to the actual geometry and cannot be regarded as neural maps in a strict sense.

Davies \etal~\cite{davies2020overfit} overfit neural networks to implicit fields of individual shapes, as a compact representation for their geometry. Implicit fields have also been used for multi-view reconstruction, where Sitzmann \etal~\cite{sitzmann_implicit_2020} optimize a neural network that represents a single shape, with the loss that favors this representation to be consistent with observed views of the object. Following the trends from \cite{vaswani_attention_2017,mildenhall_nerf_2020}, the proposed network learns to project the input into high-frequency features, hence learning in such domain. Although these techniques are similar to ours in that they use overfit networks to represent geometry, they focus on implicit surfaces and do not provide any mechanisms for inter-surface mapping. In contrast, we encode explicit surfaces via neural maps, and demonstrate that these maps can be composited and used for inter-surface mapping problems. 

Lastly, a work related to our data generation method, \cite{williams2019deep} suggest overfitting an atlas to a point cloud as a method for surface reconstruction. They however focus on the task of inferring the surface topology via overfitting, while we simply estimate a given map which already inholds the topological data with a neural network, and instead focus on the differentiability and composability properties of it, for geometry representation and for optimizing other maps composed with the parameterization.


\section{Method} 
\seclabel{method}

We now define neural surface maps and how to compute and optimize them.

\subsection{Neural Maps}
We use the term \emph{neural surface map} to refer to any neural network considered as a function $\phi:\Real^2\to\Real^n$, where the output dimension is $2$ or $3$. Indeed, this ensures the map's image is always a 3D surface, and, assuming the map is non-singular, also a $2$-manifold.

Neural surface maps can be seen as an alternative method to represent a surface map that holds two main advantages: \textit{differentiability} and ability to be \textit{composed} with other neural maps. In short, this enables us to easily compose neural maps $\phi\circ\psi$, and define an objective over the composition $o\parr{\phi\circ\psi}$ which can be differentiated and optimized via standard (deep learning) libraries and optimizers without the need to write tailor-made code to handle new objective, work with combinatorial mesh representations, or deal with the notoriously-hard map composition problem.

Furthermore, we can choose any size, architecture and activation functions for our networks, and, thanks to the universal approximation theorem~\cite{cybenko1989approximation},  know there always exists a network capable of approximating a given surface function.

We obtain and manipulate neural surface maps via two processes -- overfitting and optimization, which we detail next. 

\begin{figure}[t!]
    \centering
    \includegraphics[width=0.9\columnwidth]{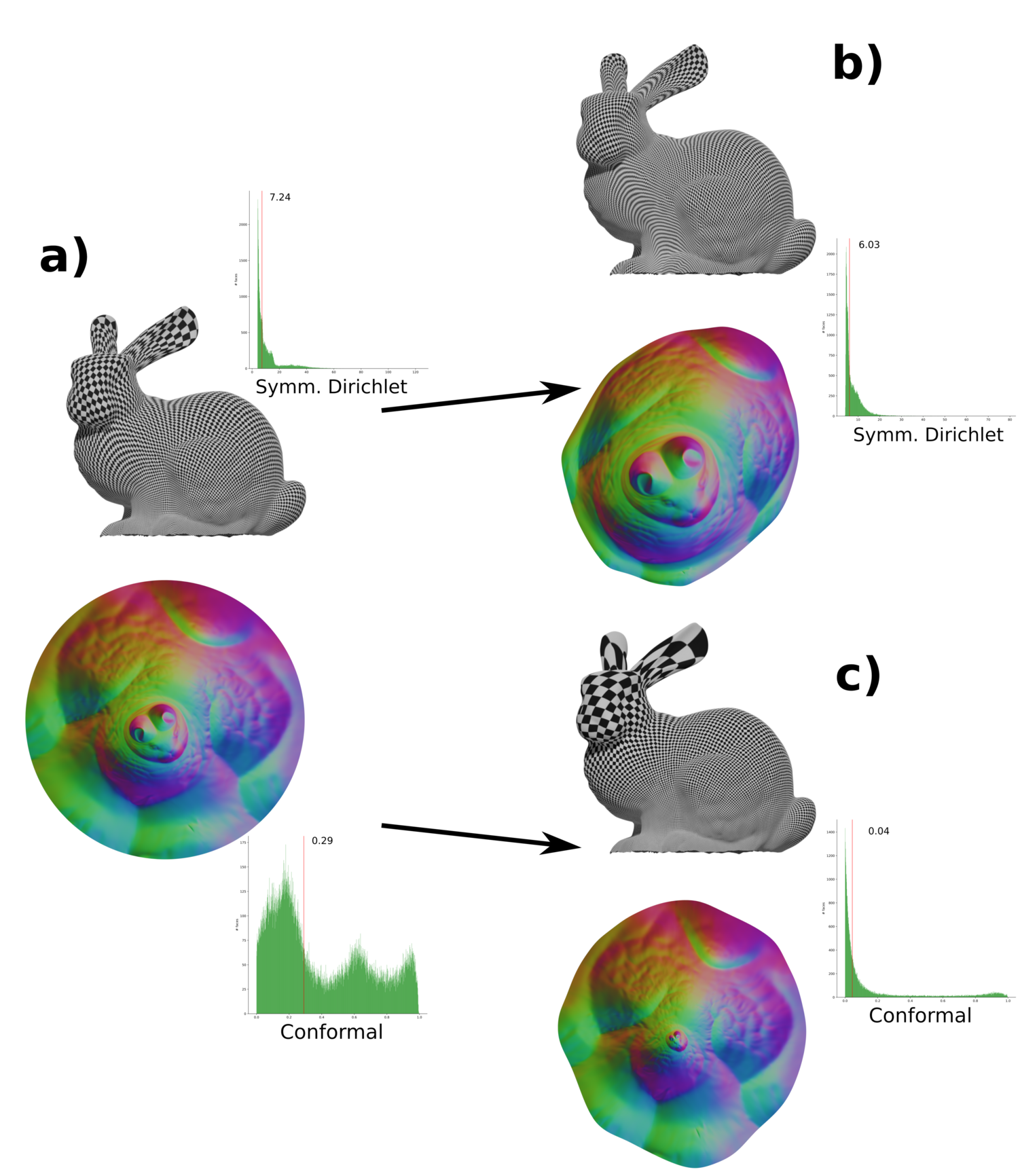}
    \caption{Free-boundary isometric~(b) and conformal~(c) parametrization of Stanford Bunny~(a) model. Independently of the size of the model, our neural maps can parametrize the input mesh, represented as neural surface map, with very few parameters. Adding a constraint over the boundary shape is as simple as regularize the mesh boundary. Initial median Dirichlet energy $D_{iso}=7.24$ is reduced to $D_{iso}=6.03$; initial median conformal energy $D_{conf}=1.29$ is reduced to $D_{conf}=1.04$.}
    
    \label{fig:coma_param}
\end{figure}
\begin{figure*}[t!]
    \centering
    \includegraphics[width=\textwidth]{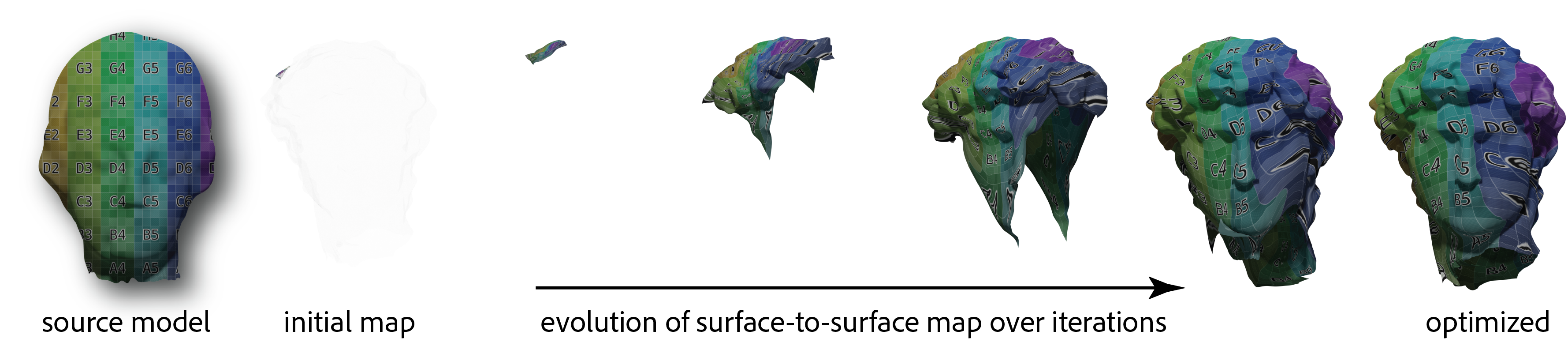}
    \caption{Evolution of surface to surface optimization between Igea and David. Final median Dirichlet energy $D_{iso}=18.25$.}
    \label{fig:surface_map_evolution}
\end{figure*}

\subsection {Overfitting Neural Surface Maps}

Let $\Omega \subset \Real^2$ be the unit circle. All our neural maps will make use of $\Omega$ as a canonical domain. Given any map $f:\Omega\to\Real^n$, we can approximate it via a neural surface map $\phi$ by using black-box methods to train the neural network and \emph{overfit} it to replicate $f$. Namely, we optimize the least-square deviation of $\phi$ from $f$ and the surface normal deviation, by minimizing the integrated error 
\begin{equation}
\label{eq:overfit_loss}
\begin{split}
    \mathcal{L}_\text{overfit} =
    \int_{p\in\Omega}\norm{f\parr{p}-\phi\parr{p}}^2 + \\
        \lambda_n \int_{p\in\Omega} \norm{n_{\phi p} - n_{fp}}^2 ,
\end{split}
\end{equation}
where $n_{\phi p}$ is the estimated normal at $p$, and $n_{fp}$ is the ground truth normal.
In case $f$ is indeed a continuous map, such as a piecewise-linear map mapping triangles to triangles, we can optimize this objective by approximating the integral in Monte-Carlo fashion by summing the integrand over a random set of sample points.  
Namely, to use neural surface maps to represent surfaces, we first compute a ground truth map $f$ by overfiting to a UV parameterization of the mesh into 2D, computed via any bijective parameterization algorithm of our choosing -- in this paper, we show results with SLIM~\cite{rabinovich2017scalable}, by which we achieve an injective map of the mesh into $\Omega\subset \Real^2$. We consider the inverse of this map, which maps $\Omega$ back into the 3D mesh $\mathcal{S}$, as our input $f:\Omega\to\mathcal{S}$, and overfit $\phi$ to it by minimizing Equation~\ref{eq:overfit_loss}. Thus, we obtain a neural representation of the surface. More specifically, this is a \emph{mapping} into the surface, endowed with specific UV coordinates, with point $\phi\parr{x,y}$ having UV coordinates $x,y$. 
\figref{fig:bimba_nueral_map} shows several examples of such  overfitted neural maps and their faithfulness to the original geometry. Our method can faithfully represent smooth shapes as well as those having sharp edges. Note that we assume that the objects are or have been cut open to be homeomorphic to a disc. 

Before progressing to discussing how can we compose maps and optimize them, we define the distortion measures we wish to optimize.

\subsection{Surface Map Distortion}
We wish to optimize several energies related to neural surface maps. Similarly to \cite{bednarik_shape_2019}, for a neural map $\phi:\Omega\to\Real^n$, we denote by $J_p\phi\in\Real^{n\times2}$ the matrix of partial derivatives at point $p\in\Omega$, called the \emph{Jacobian} of $\phi$. The Jacobian essentially quantifies the local deformation at a point. Letting $M_p=J_p^TJ_p$, we subsequently  can quantify the symmetric Dirichlet energy \cite{rabinovich2017scalable},
\begin{equation}
 D_\text{iso} = \int_\Omega\text{trace}\parr{M_p}+\text{trace}\parr{\parr{M_p+\varepsilon I}^{-1}}
 \label{eq:dirichlet}
\end{equation}
where $I$ is the identity matrix, added with a small constant $\epsilon$, set to $0.01$, to regularize the inverse. 

Likewise, we can define a measure of conformal distortion via 
\begin{equation} \label{eq:conf_loss}
 D_\text{conf} = \int_\Omega\norm{\frac{\text{trace}\parr{M_p}}{\norm{M_p}^2}M_p-I}^2.
\end{equation}
We evaluate the integrals by random sampling of the function in the domain. 

Next, we show how to define surface-to-surface maps via various compositions of the maps and optimize their distortion, in the pairwise and in the shape collection setting.

\subsection{Geometry-preserving optimization via composition}

Our basic representation of 3D geometries is, as discussed above, via an \emph{overfitted} neural surface map $\phi:\Omega\to\Real^3$ that approximates a given map $f$. We now treat  $\phi$ as our de-facto representation of the geometry. Our goal is to optimize various properties relating to the surface map, without affecting the geometry. However, optimization of the map is not trivial since it will immediately change our 3D geometry. We propose a solution to completely avoid this issue, next. 

Assume we are given a neural surface map representing some surface $\phi:\Omega\to\mathcal{S}$; we wish to optimize the distortion $D\parr{\phi}$ of the map. It is immediate to optimize $\phi$ \emph{itself} with respect to our differentiable notion of distortion, however that will cause the map to change, and thus its image, the 3D surface, will change and could, for instance, flatten to the plane. To overcome this,  we suggest introducing another neural surface map $h: \Omega\to\Omega$. We can now define a new map, $\phi^h = \phi\circ h$. As long as we solely optimize $h$ and ensure it is onto $\Omega$, we are guaranteed that the image of $\phi^h$ is still the original image of $\phi$, \ie, respects the original surface. 

We can now optimize the distortion of $\phi^h$, by optimizing $h$ and keeping $\phi$ fixed, thereby finding a map from $\Omega$ to $\mathcal{S}$ which is (at least a local) minimizer of the distortion measure of our choice:
$$
\min_h D\parr{\phi^h} .
$$
The distortion is a differentiable property of the map and hence is readily available, \eg, via automatic differentiation. In fact, composition, and minimization of distortion can be achieved in a mere few lines of code in Pytorch.

We can now consider composing more than two of these maps, to enable maps into more intricate domains.

\subsection{Compositing Neural Maps} 
\label{ss:compose}
\paragraph{Map composition via common domains.} One of the many advantages of our representation's composability is to enable representing maps between a pair of surfaces, using the classic method of a common domain, as depicted in Figure \ref{fig:teaser}: we posses two overfitted neural maps, $\phi,\psi:\Omega\to\Real^3$, respectively representing two surfaces $\mathcal{S},\mathcal{T}$, and we wish to define and optimize an inter-surface mapping between these two 3D surfaces, $f:\mathcal{T}\to\mathcal{S}$.

\begin{wrapfigure}[7]{r}{0.45\columnwidth}
\vspace{-26pt}
\hspace{-27pt}
\includegraphics[width=0.55\columnwidth]{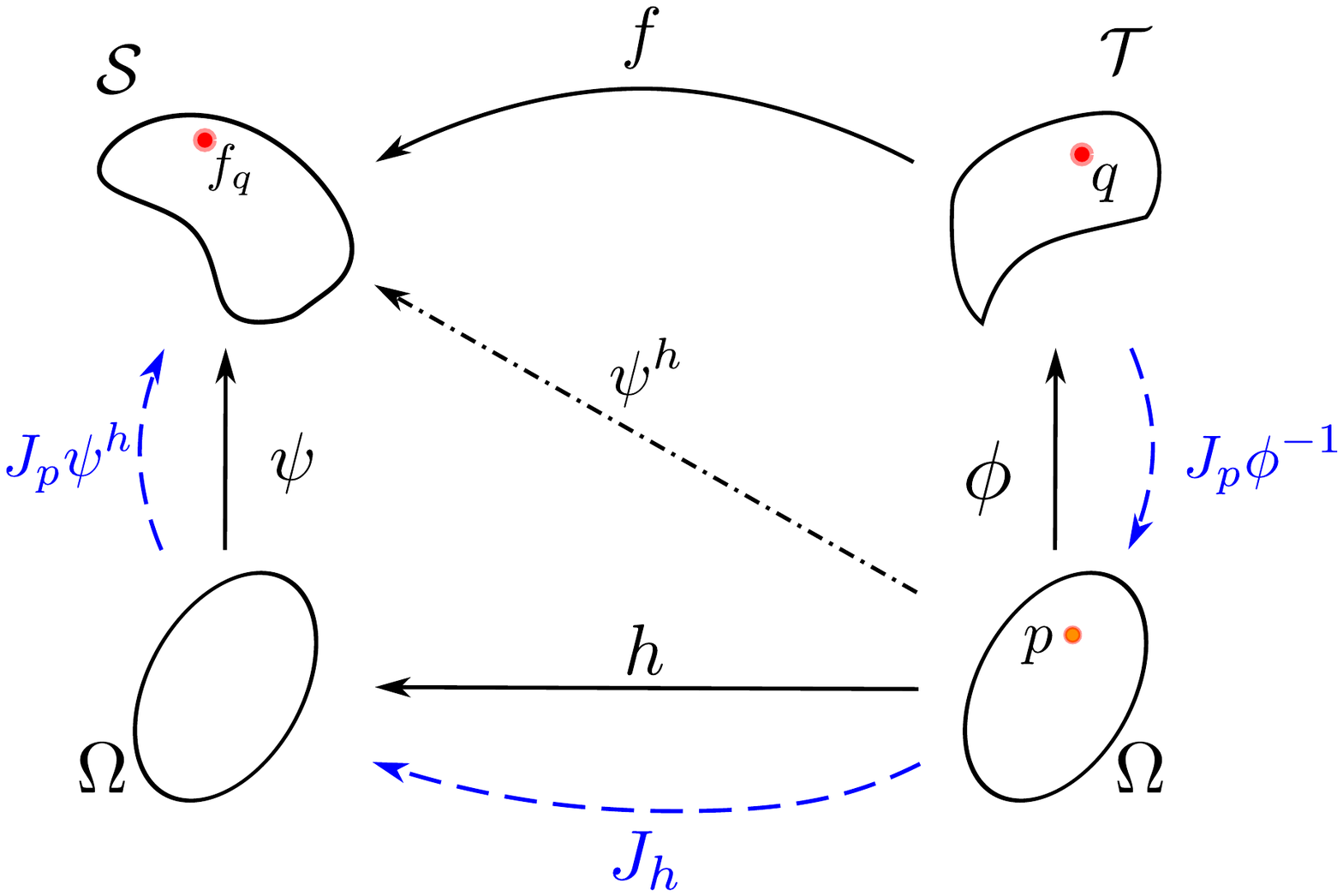}
\vspace{0pt}
\end{wrapfigure} 
To address the above problem, we define as before a map  $h:\Omega\to\Omega$ and the composition $\psi^h = \psi \circ h$. At first glance, it would seem that in this case, to map a point from $\mathcal{T}$ to $\mathcal{S}$, we will need to consider the map $\psi^h \circ \phi^{-1}$, which includes an inverse of the \emph{entire} map, that is of course not readily tractable.

However, we can define the map $f$ via the following simple definition: for any point $p\in\Omega$, $f$ is implicitly defined as the map satisfying $f\circ\phi \triangleq \psi^h$, or in simple words: for any point $p\in\Omega$, $f$ matches the image of $p$ under $\phi^h$ with the image of $p$, mapped through $h$ and then through $\psi$ (refer to Figure \ref{fig:teaser} for an illustration). This definition is known as the \textit{common domain} definition of a map and has been  used in many works \cite{Lee:1999,Schreiner:2004:IM:1015706.1015812,Kraevoy:2004:CCR:1015706.1015811,Bradley08, aigerman2014,weber2014locally,aigerman2015,aigerman_hyperbolic_2016}. It is easy to verify that this definition is identical to the one using the inverse, as long as the inverse exists, and can still provide a bijective map between the surfaces even in cases where it does not exist (cf., \cite{weber2014locally,aigerman_lifted_2014}).

\paragraph{Computing distortion in the common domain.} Even though $f$ itself is not tangible for optimization, as it is implicitly defined by $h$, luckily the only differential quantity we need from $f$ to compute the distortion, is the Jacobian of $f$, denoted $J_qf$ at point $q=\phi\parr{p}$. Using basic differential calculus arithmetic, $J_qf$ can be derived to be exactly 
\begin{equation}
\label{eq:jac_comp}
    J_qf = J_p{\psi^h}\parr{J_p\phi}^{-1},
\end{equation}
which is composed of the Jacobian of $\psi$ and the inverted Jacobian of $\phi$ at point $p$, both readily available. 
Hence to optimize the distortion of $f$, we can take \eqref{eq:jac_comp}, and plug it as the Jacobian  used to define $M$ in one of the distortion measures \eqref{eq:dirichlet},\eqref{eq:conf_loss}, which we denote as $D\parr{f}$.

\paragraph{ Optimizing $h$ for bijectivity. } In order for $h$ to indeed be a well-define surface map, it needs to map exactly bijectively (\ie, 1-to-1 and onto) to the source domain of $\psi$, which is $\Omega$. To ensure that, we only need to ensure that $h$ has a positive-determinant Jacobian everywhere, and maps to the target boundary injectively. We optimize $h$ to map the boundary onto itself, via the energy
\begin{equation}
 B\parr{h} = \int_{p \in \partial \Omega}\sigma\parr{h\parr{p}},
\end{equation}
where $\sigma$ is the squared signed distance function to the boundary of $\Omega$. Note that the boundary map is free to slide along the boundary of $\Omega$ during optimization, enabling the boundary map to change. This is true for all points on the boundary, except those mapped to the four corners which are fixed to place and are essentially keypoint constraints between the two models.

Further, we also optimize $h$ to encourage its Jacobian's determinant to be positive, via
\begin{equation}
     G =\lambda_\text{inv}\int \max \parr{- sign\parr{\abs{J_h}} exp\parr{-\abs{J_h}}, 0}.
\end{equation}

\paragraph{Keypoint constraints.} Lastly, in many cases, a sparse set of corresponding key points on the two surfaces are given, and it is required that the surface map $f$ maps those points to one another. Given keypoints on $\mathcal{S}$, we can, in a preprocess before optimization, find their preimages in $\Omega$, to get a set of points $P$ s.t. $\phi\parr{P_i}$ maps to the $i$th keypoint. We likewise can find the preimages of the keypoints from $\mathcal{T}$ and their preimages $Q$ under $\psi$. If these key points are required to be mapped to one another between the two surfaces by $f$, we can achieve that by requiring $h\parr{P_i}=Q_i$, which guarantees the induced $f$ maps the points correctly. We optimize for that equality by reducing its least-squares error:
\begin{equation}
    C\parr{h} = \lambda_C\sum_i\norm{h\parr{P_i} - Q_i}_2^2 .
\end{equation} 
To facilitate the optimization, we apply a rotation, $R$, to the input of $h$. $R$ is pre-computed from the landmarks.

\paragraph{Optimization for surface-to-surface maps.} To compute the surface map, we optimize the distortion of $f$ with respect to $h$, while ensuring $h$ respects the mapping constraints
\begin{equation}
    \min_h D\parr{f} + C\parr{h} + B\parr{h}+G\parr{h}.
\end{equation}
This yields a map $h$ that maps onto the domain square,  and represents a distortion-minimizing surface map $f$ that maps the given sets of corresponding keypoints correctly, as shown for instance in Figure \ref{fig:teaser}.

\paragraph{Cycle-consistent surface mapping.} We also extend our method to discover inter-surface mapping among a collection of  $k$ surfaces $\mathcal{S}_1,\mathcal{S}_2,...,\mathcal{S}_k$ represented respectively via neural maps $\phi_1,\phi_2,...,\phi_k$, we can define a \emph{cycle consistent} \cite{nguyen2011optimization,huang2013consistent}
set of surface maps by considering $k$ additional neural maps, $h_i:\Omega\to\Omega$, define the composition $\phi^h_i = \phi_i\circ h_i$, and then define the surface-to-surface maps $F_{i\to j}:\mathcal{S}_i\to\mathcal{S}_j$ via $F_{i\to j}\circ\phi^h_i \triangleq \phi^h_j$. This naturally allows extracting a set of mutually consistent maps while additionally optimizing for (all pairs) surface-to-surface maps, see Figure \ref{fig:collection_map}. Note that achieving similar qualities via classic methods is significantly challenging, and to the best of our knowledge, while previous methods could compute cycle consistency, none could optimize for true surface-surface distortion minimization over the entire collection.

\section{Experiments} \seclabel{experiments}
We use our neural-mapping representation in the context of various mapping problems, such as surface parameterization, inter-surface mapping, and mapping a collection of shapes. See the supplementary material for more examples. \\

\noindent \textbf{Neural Mapping.}
For all surfaces shown in this paper, we render the reconstructions obtained with our neural mapping representation. Note how our overfitting procedure is able to capture even very detailed features of the original shape with a high fidelity. Figure~\ref{fig:bimba_nueral_map} illustrates the difference between our reconstruction and the input mesh (highlighted in red).  There are minor discrepancies between the models in regions like hairs of the bust and paws of the rhino. We observe that our reconstructions tend to be slightly smoother than the original shapes due to the use of softplus. \\
\begin{figure}[t]
    \centering
    \includegraphics[width=\columnwidth]{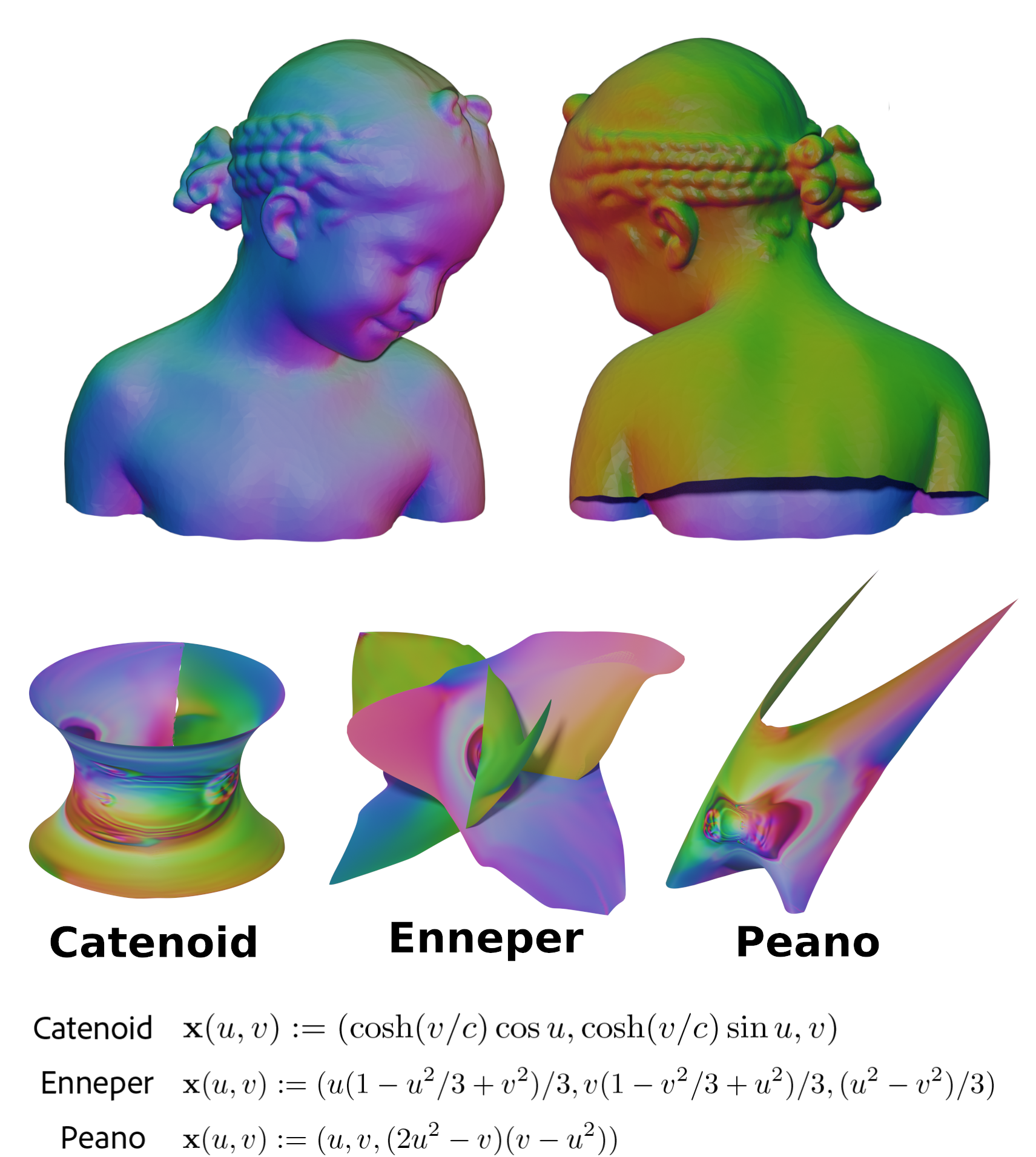}
    \caption{Surface maps between a  neural surface mapping representing the Bimba model, into several surfaces represented directly via analytic functions. Colors are based on the normals of Bimba model. Please refer tot he supplemental for further visualizations. }
    \label{fig:analytic_map}
\end{figure}

\noindent \textbf{Surface Parameterization.}
The main advantage of neural mapping is not in representing the surfaces, but in representing the mapping. We now take the map  $\phi:\Omega\to\Real^3$ from Figure~\ref{fig:bimba_nueral_map}, and introduce another map $h:\Omega\to\Real^2$, where we don't constrain its output domain. Similarly to the discussion in Subsection \ref{ss:compose}, we can define the map $f$ from the 3D model implicitly via as $f\parr{\phi\parr{p}}=h\parr{p}$ for all $p\in\Omega$. We then minimize the isometry distortion of $f$ (Eq.~\ref{eq:dirichlet}), using the method to extract the Jacobian discussed in Subsection \ref{ss:compose}.  Note that this objective is different from the one that was used to produce $\phi$, hence we undo the original parameterization's distortion by compositing the neural map with a newly optimized map in Figure~\ref{fig:coma_param}. See the supplementary for more results.

In contrast to UV parameterizations of meshes, the complexity of our optimization for this composition is completely independent of the resolution of the geometry. \\

\noindent \textbf{Surface-to-surface Maps.}
We can obtain a surface-to-surface map by compositing neural maps with a map between two atlases, as discussed in Subsection \ref{ss:compose}. In Figure~\ref{fig:surface_map_evolution}, we show the evolution of the map during optimization.  Note how despite significant geometric differences between surfaces, the result is a bijective, low-distortion mapping. Please see more such maps in the supplementary. \\

\noindent \textbf{Composition with Analytical Maps.} Our method can optimize an inter-surface map $f$ from $\phi,\psi$ just as well when $\psi$ is not a neural map, but rather an analytical mapping defining some surface. Indeed, only $h$ itself is required to be neural in our formulation of surface-to-surface maps. In  Figure~\ref{fig:analytic_map}, we show mappings of Bimba into three such analytical surfaces. In this case, we optimize the conformal distortion \eqref{eq:conf_loss} of $f$. Please refer to the supplementary for further visualizations. \\

\begin{figure}[t]
    \centering
    \includegraphics[width=0.9\columnwidth]{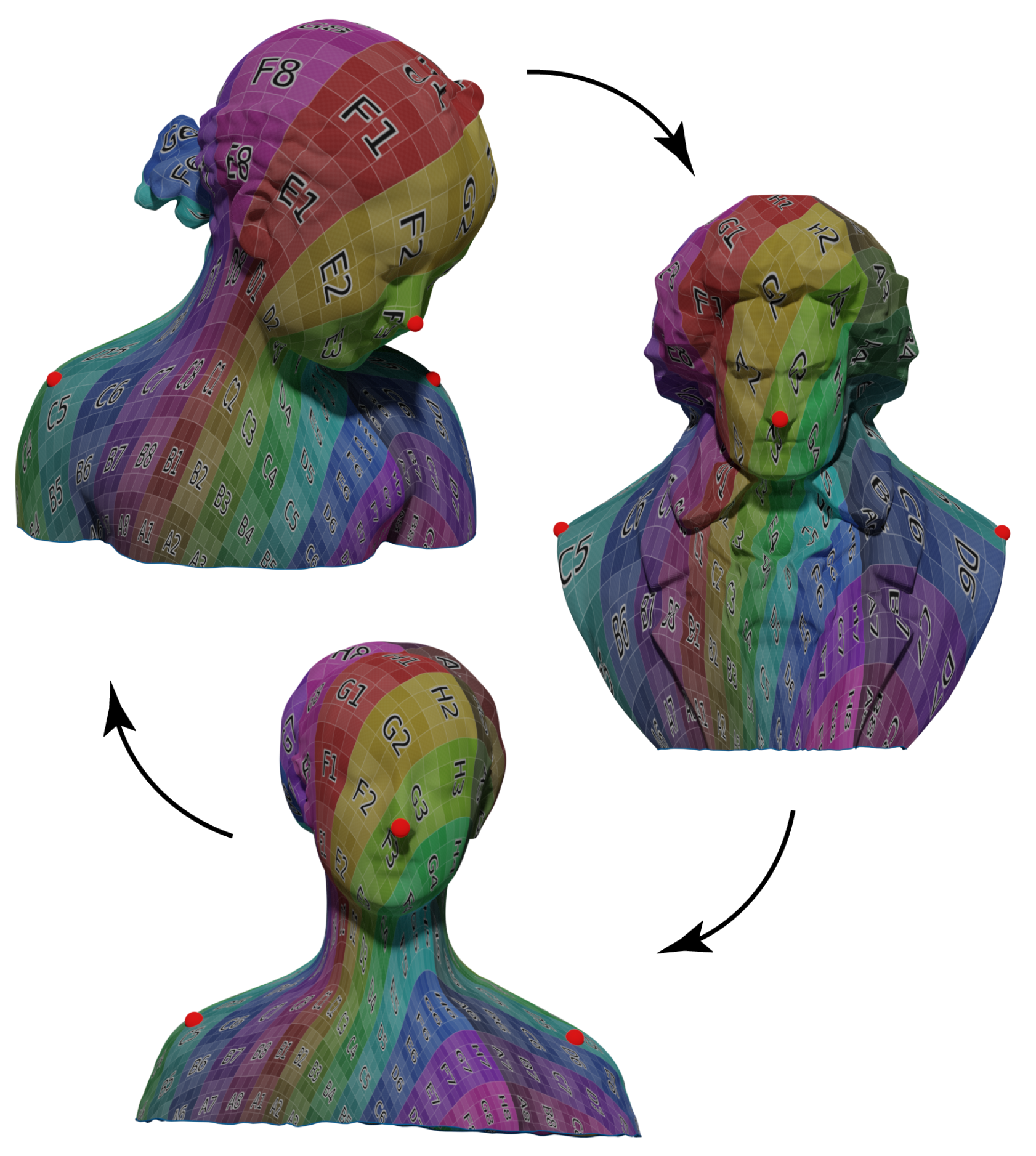}
    \caption{Collection mapping. We map from one model to the other through a neural map. We then minimize the distortion between each different model. Cycle consistency is ensured by construction.}
    \label{fig:collection_map}
\end{figure}

\noindent \textbf{Cycle-consistent Mapping for Collections of Surfaces.}
Finally, we show that thanks to the compose-ability of neural surface maps, our method can be efficiently applied to cycle-consistent mapping problem for a collection of shapes. Furthermore, since we use a common domain, the maps are guaranteed to be cycle-consistent, as in \cite{nguyen2011optimization,huang2013consistent}. We minimize the isometric distortion of the surface-to-surface maps between all pairs of surfaces in a collection of three models, following the method discussed in Subsection \ref{ss:compose}. Figure~\ref{fig:collection_map} illustrates that we were able to obtain cycle-consistent low distortion maps between all shapes in the collection. We used one keypoint on the nose and shoulders of each model to ensure correct alignment. See the supplementary for more collection-maps.

\begin{figure}[t]
    \centering
    \begin{tabular}{ccc}
         \includegraphics[width=0.23\columnwidth]{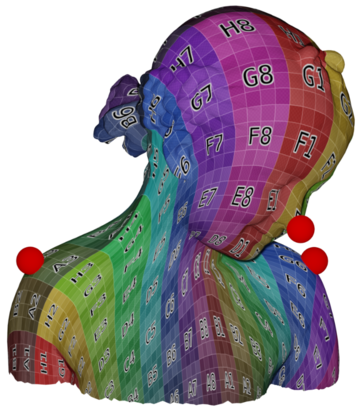} &
         \includegraphics[width=0.23\columnwidth]{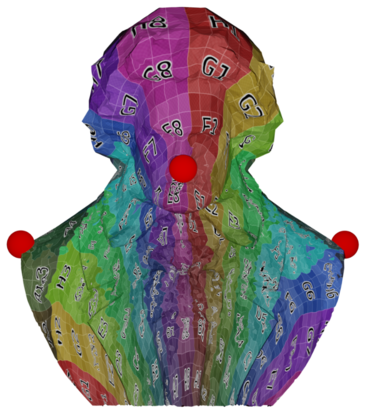} &
         \includegraphics[width=0.23\columnwidth]{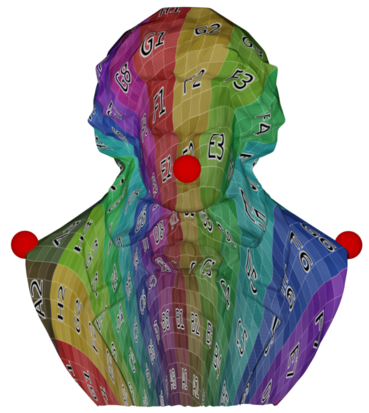} \\
         
         \includegraphics[width=0.28\columnwidth]{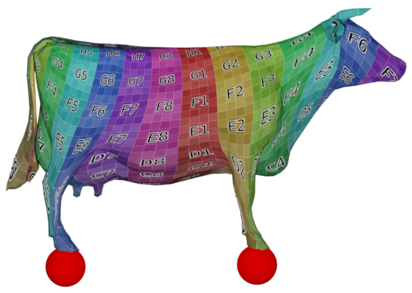} &
         \includegraphics[width=0.28\columnwidth]{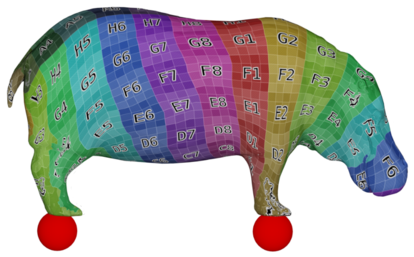} &
         \includegraphics[width=0.28\columnwidth]{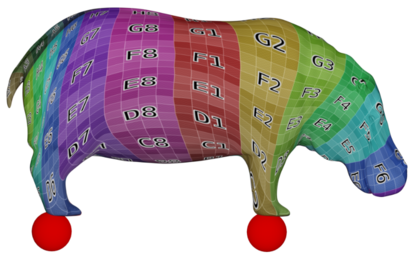} \\
         
         \textbf{(a)} source &
         \textbf{(b)} \cite{Schreiner:2004:IM:1015706.1015812} &
         \textbf{(c)} ours \\
    \end{tabular}
    \caption{Comparison with inter-surface mapping~\cite{Schreiner:2004:IM:1015706.1015812}. 
    }
    \label{fig:compare_classic}
\end{figure}

\begin{figure}[t]
    \centering
    \begin{tabular}{ccc}
         \includegraphics[width=0.23\columnwidth]{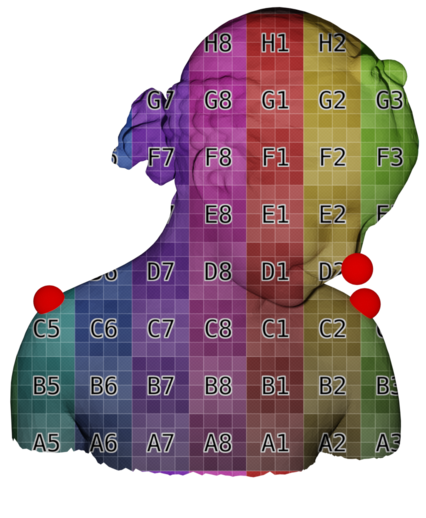} &
         \includegraphics[width=0.23\columnwidth]{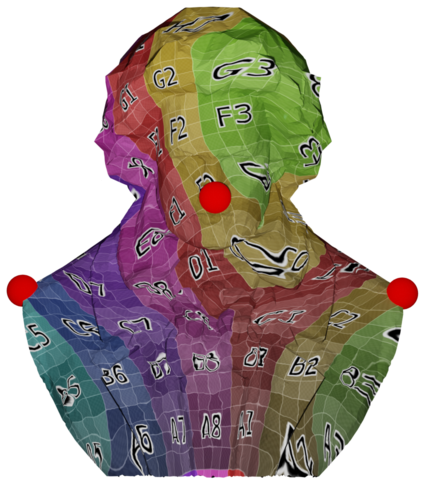} &
         \includegraphics[width=0.23\columnwidth]{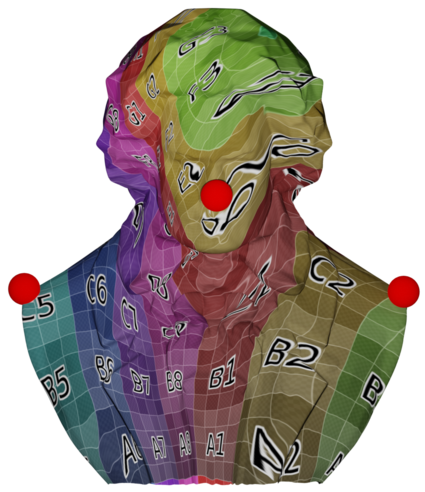} \\
         
         \includegraphics[width=0.28\columnwidth]{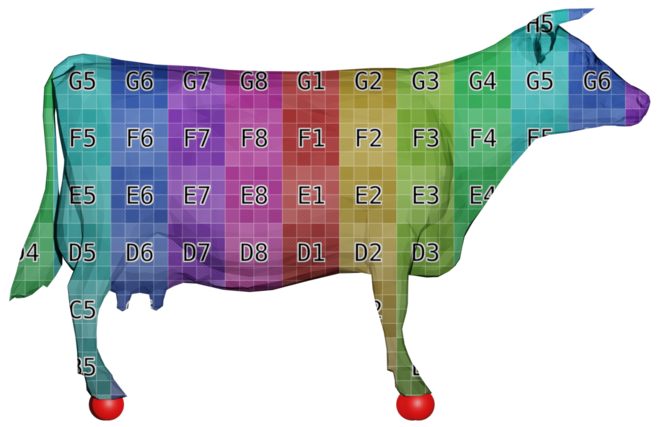} &
         \includegraphics[width=0.28\columnwidth]{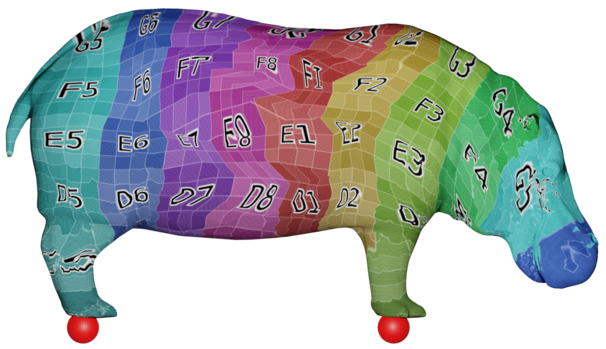} &
         \includegraphics[width=0.28\columnwidth]{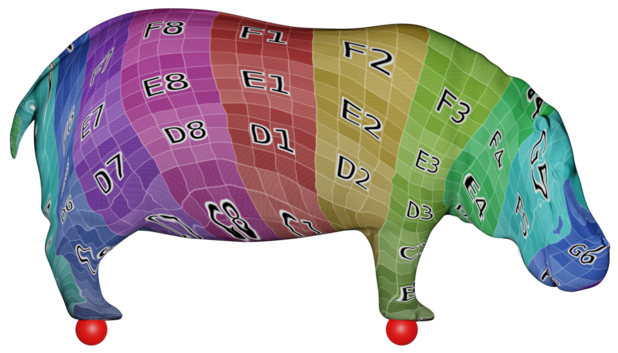} \\
    
         \textbf{(a)} source &
         \textbf{(b)} \cite{mandad2017variance} &
         \textbf{(c)} ours \\
    \end{tabular}
    \caption{Comparison with state of art shape correspondence~\cite{mandad2017variance}.}
    \label{fig:compare_sota}
\end{figure}

\noindent \textbf{Baseline comparison.}
To validate neural surface maps, we offer visual comparisons with the classic inter-surface method~\cite{Schreiner:2004:IM:1015706.1015812} and Mandad \etal~\cite{mandad2017variance}. Schreiner \etal~fails to produce smooth maps while matching landmarks: respectively for the bust and animal shown in \figref{fig:compare_classic}, \cite{Schreiner:2004:IM:1015706.1015812} presents $8.58\%$ and $8.54\%$ triangles flips with a median $D_{iso}=4.90$ and $D_{iso}=7.00$. Similarly, Mandad \etal~achieve a median $D_{iso}=7146$, with $49.91\%$ of flips, and $D_{iso}=10669$ with $49.86\%$ of flips, see \figref{fig:compare_sota}. Note, \cite{mandad2017variance} introduces discontinuities in the map, resulting in large distortion and misalignment. On the other hand, our method offer a continuous, properly aligned, map. Numerically, our map for busts exhibit $D_{iso}=7.00$ with no triangle flips, $D_{iso}=8.56$ and $0.03\%$ flips for animal.

\subsection{Implementation Details}

In all our experiments, we use a neural network consisting of ten-layer residual fully-connected network, with 256 hidden units per layer, with a Softplus activation function. We use $\lambda_n=0.01$, $\lambda_B=10^{6}$, $\lambda_{inv}=10^{2}$, $\lambda_C=10^{3}$ in all experiments. 
We sample  the initial mesh uniformly with 500k points. Since our goal is to fully-optimize the networks, they are trained until the gradient's norm drops below a threshold of $0.1$. In all cases, we optimize the network with and RMSProp, and initialize the optimization procedure with a learning rate of $10^{-4}$ and momentum 0.9, the step size is modulated with \cite{loshchilov2016sgdr}. Similarly, maps used for surface mapping are four-layer fully-connected network of 128 hidden units, with Softplus. In general, overfitted networks converge in 3-7h based on the complexity of the model, while, surface-map and collection-map optimization take around 3h to reach a stable configuration.

\section{Conclusions and Future Works} 
\seclabel{conclusion}

We introduced neural surface maps as a core representation for surfaces that is easily differentiable and compose-able. Using the common domain approach, we can easily use these traits to optimize for different properties. Overfit to individual meshes allows encoding shapes as network weights, and subsequently optimize maps while keeping the surface approximation quality fixed. We demonstrated the universality of neural maps addressing a wide range of challenging classical tasks including parameterization, surface-to-surface distortion minimization, and extracting maps across a collection of shapes.

Our work has several limitations. For one, we only discussed representing disk-topology surfaces. Other topologies can be approached with cuts. Second, we relied on the assumption of $h$ being bijective and mapping the keypoints correctly; in theory, we cannot guarantee that this requirement is upheld, however, in our experiments, it is rare for this condition to be violated.

We see many immediate uses to the differentiability and composability of our representation, such as applying differential geometry operators to the models as well as solving PDEs on them. Resorting to neural network generalization capabilities can bring large high-resolution dataset within our reach, exposing neural surface maps to applications like segmentation and classification.

\section*{Acknowledgements}
LM thanks Manish Mandad for helping comparing with \cite{mandad2017variance}. LM was partially supported by the UCL Centre for AI and the UCL Adobe PhD program.

{\small
\bibliographystyle{ieee_fullname}
\bibliography{references}
}

\cleardoublepage

\beginsupplement

\section{Supplementary material}

\subsection{Surface-to-surface Maps}
As described in the paper, we can represent and optimize surface-to-surface maps by composing two neural atlases $\phi$ and $\psi$ with $h$ that maps one atlas to the other.

\begin{figure}[t]
    \centering
    \includegraphics[width=1.0\columnwidth]{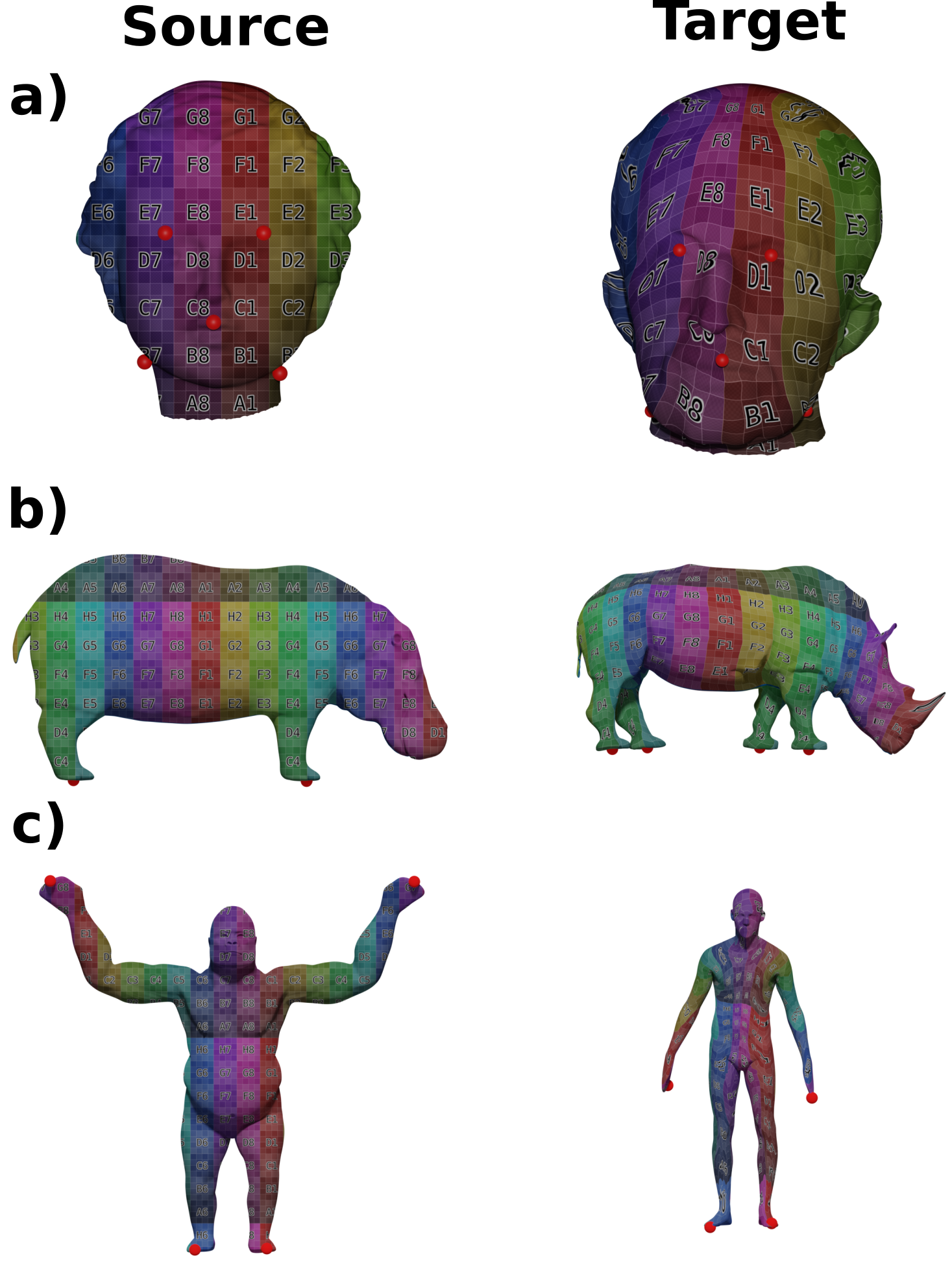}
    \caption{Surface to Surface maps. Red spheres depicts landmarks.}
    \label{fig:s2s_maps}
\end{figure}

\figref{fig:s2s_maps} depicts more examples of surface maps.  In all cases we use landmarks to guide the distortion to the correct minimum.

\subsection{Cycle-consistent Mapping for Collections of Surfaces}

\begin{figure}[t]
    \centering
    \begin{tabular}{c}
        \includegraphics[width=0.9\columnwidth]{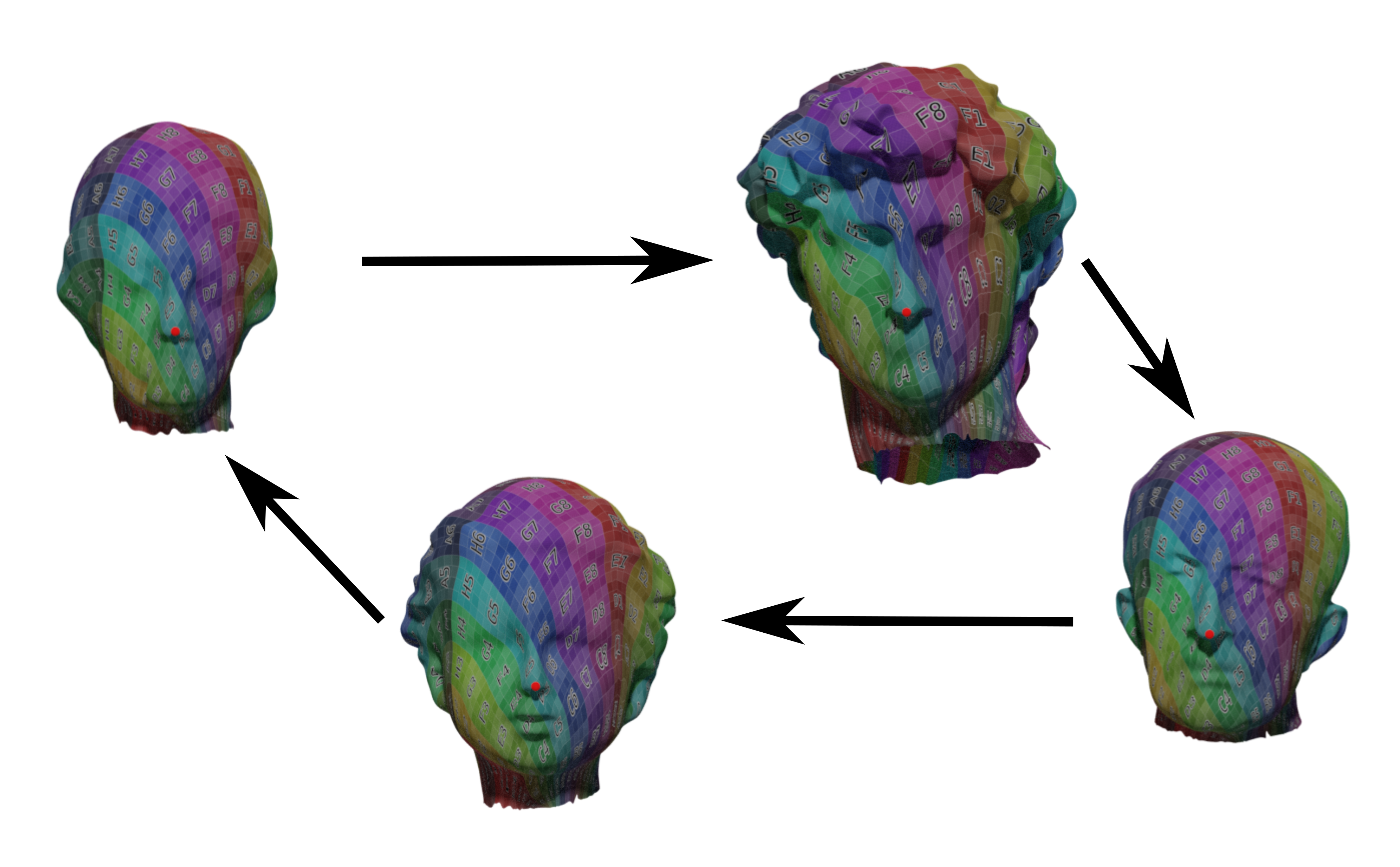} \\
        \textbf{(a)} \\
        \includegraphics[width=0.9\columnwidth]{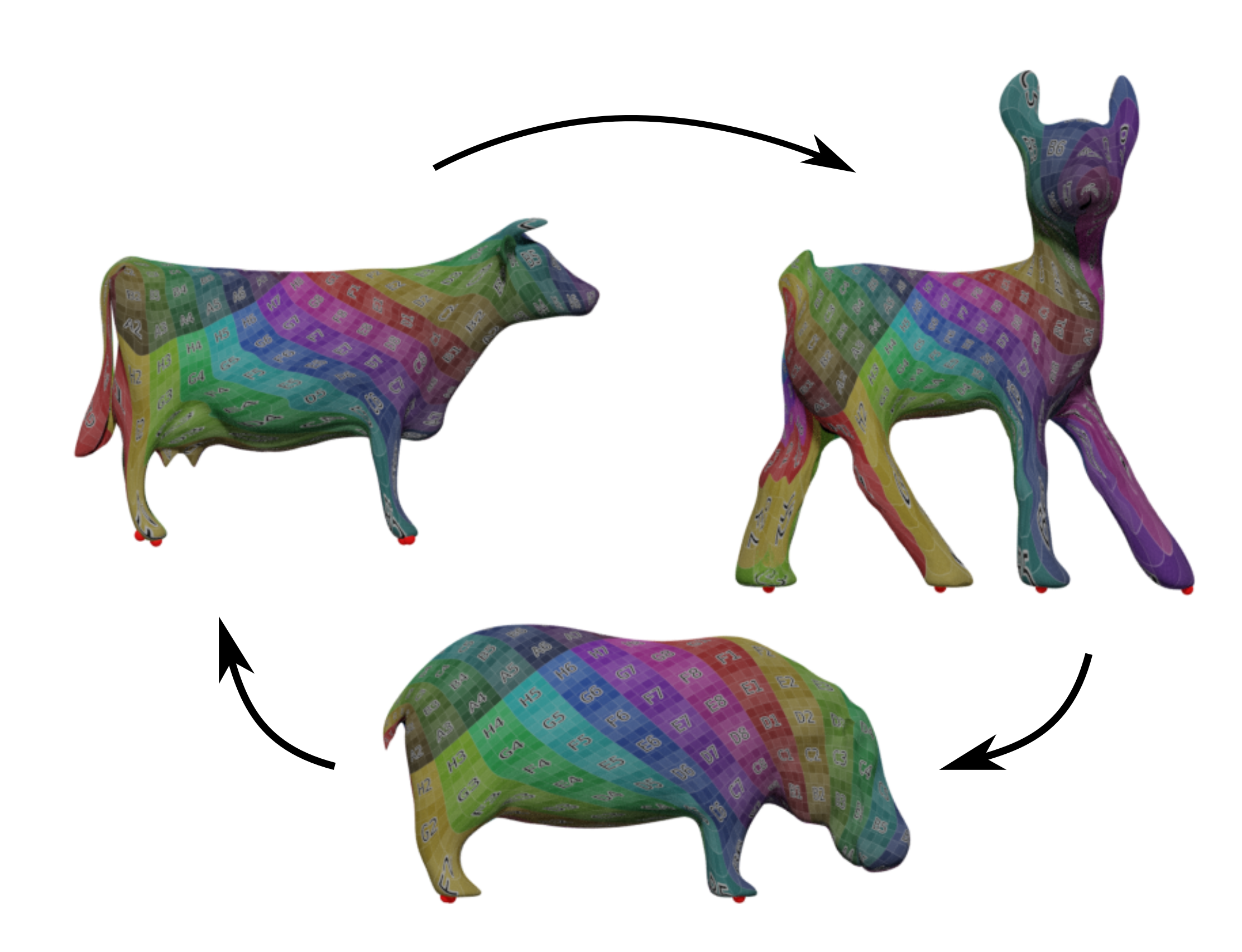} \\
        \textbf{(b)} \\
    \end{tabular}
    
    \caption{Cycle-consistent maps over a collection. We minimize the isometric distortion of the maps between all pairs of shapes. Landmarks are visualized as red sphere.}
    \label{fig:coll_heads}

\end{figure}

As discussed in the paper, thanks to our formulation it is possible to optimize all maps between a collection of shapes. As previously illustrated, a collection of neural maps is inherently cycle-consistent, and we can, for the first time, optimize the distortion of all maps in the collection. \figref{fig:coll_heads} depicts surface maps over a collection of heads (a) and a 4-legged models (b).  In both cases we employ landmarks highlighted as red spheres. 

\begin{figure*}[t]
    \centering
    \begin{tabular}{ccc}
        \includegraphics[width=0.6\columnwidth]{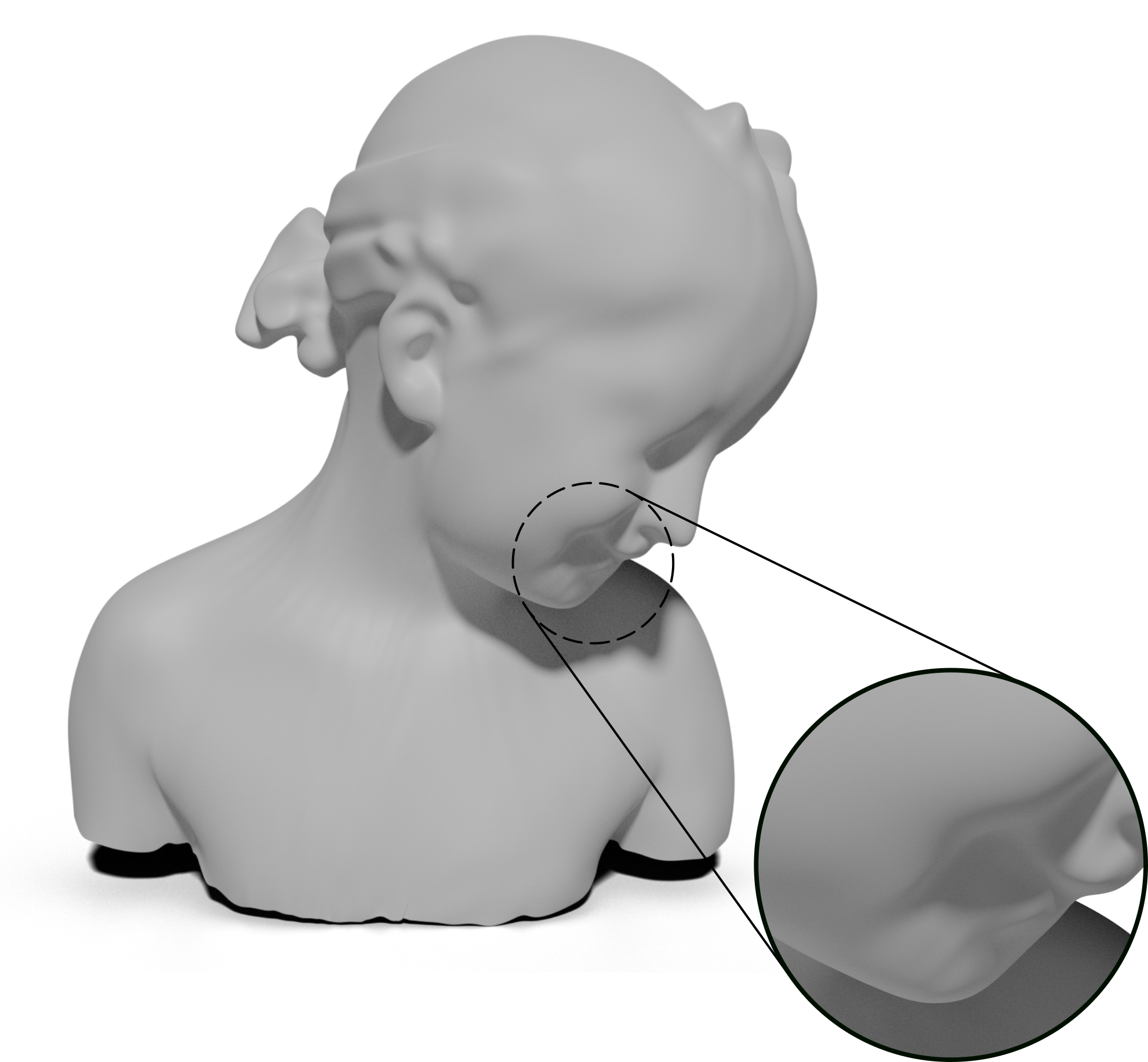} & 
        \includegraphics[width=0.6\columnwidth]{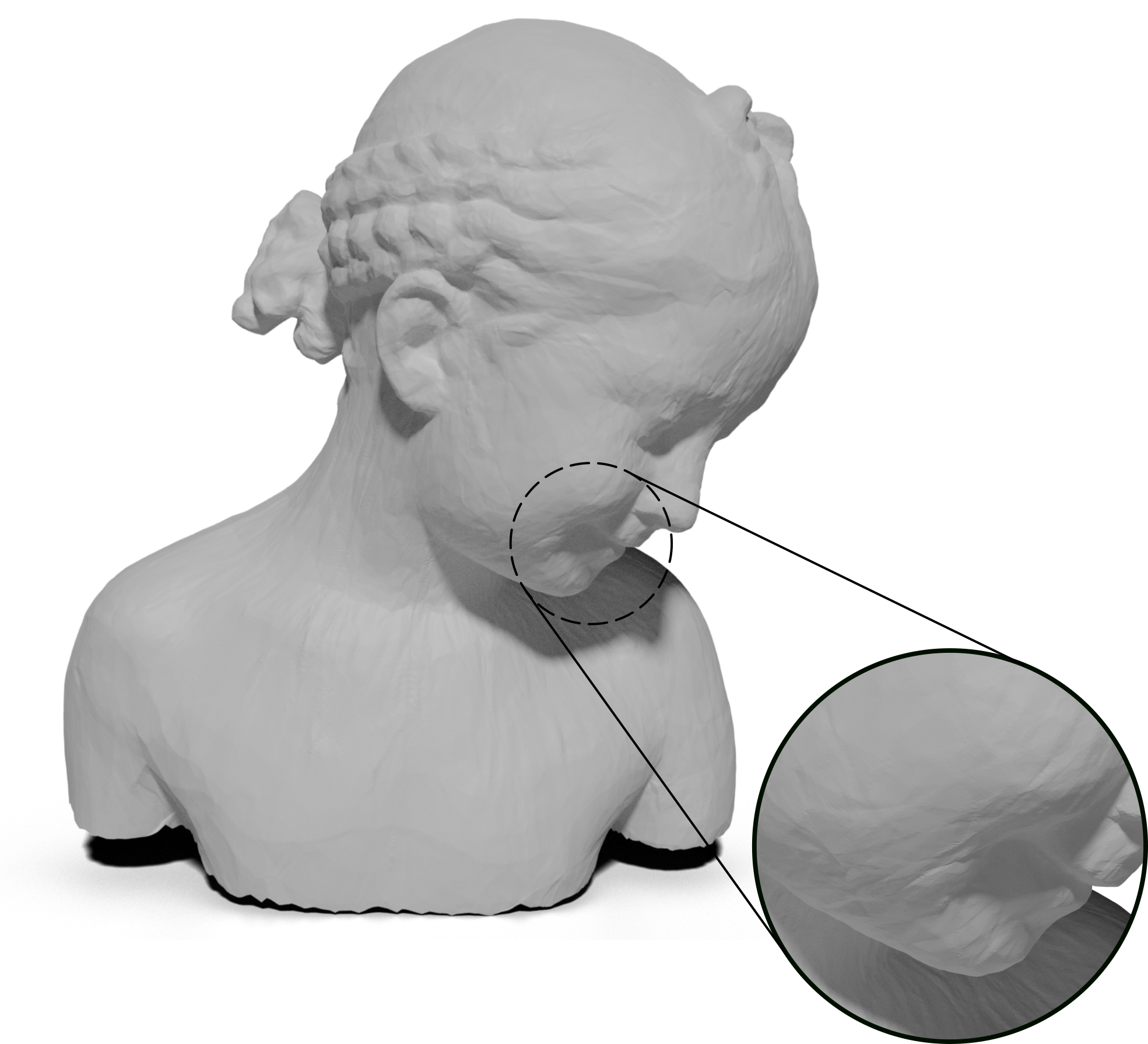} &
        \includegraphics[width=0.6\columnwidth]{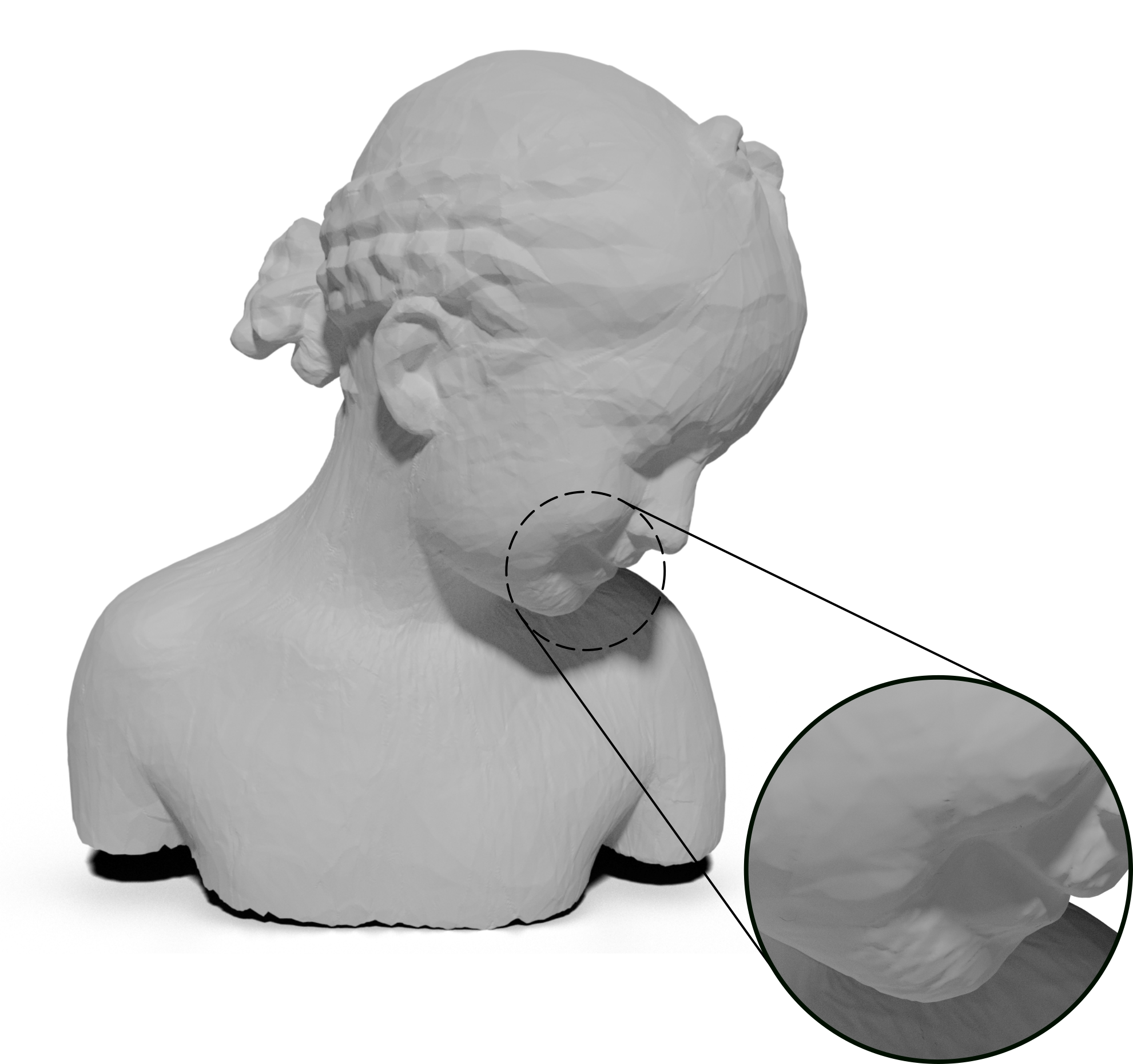} \\
        \textbf{(a)} & \textbf{(b)} & \textbf{(c)} \\
    \end{tabular}
    
    \caption{A MLP-based neural surface map with different activation functions. Softplus (a) maps are smooth. LeakyRelu (b) and Relu (c) are more oscillatory.}
    
    \label{fig:overfit_act}
\end{figure*}

\subsection{Approximating Surfaces}

\begin{figure*}[t]
    \centering
    \includegraphics[width=1.8\columnwidth]{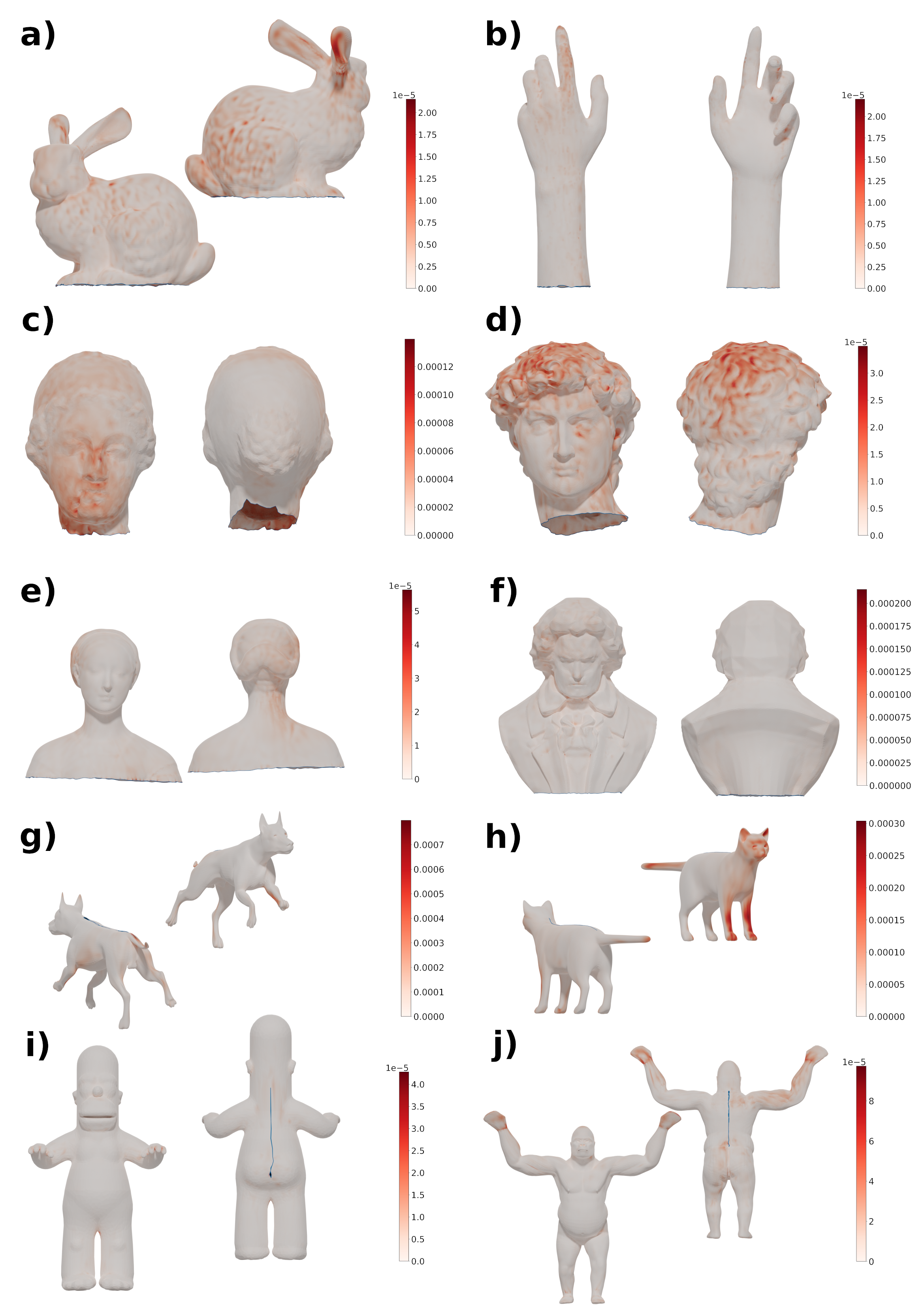}
    \caption{Neural surface maps depicted are color-coded to highlight the error compared to the ground truth mesh. In general, the produced models have low deviation from the original surface.}
    \label{fig:overfit_suppl}
\end{figure*}

As described in the main paper, a \emph{neural map} $\psi$ can approximate a given surface $S$ by approximating the atlas $f:\Real^2\to S$ accurately. To further demonstrate the effectiveness of our formulation, in \figref{fig:overfit_suppl} we show more overfitted models for categories such as Stanford Bunny~(a), hand~(b), heads~(c to d), busts~(e and f), animals~(g and h), and human~(i and j). Each neural surface map consists of a ten-layer residual fully-connected network with Softplus activation.

In \figref{fig:overfit_act} we compare different activation functions such as LeakyRelu and Relu. These non-smooth functions introduce artefacts, such as unwanted wrinkles \figref{fig:overfit_act}~(b and c). Overlooking this behavior might bear negative effects in map composition as these introduced details can be amplified or inject distortion in the final map. The use Softplus alleviates these artefacts, but biases the map towards smooth surfaces, hence the hairs in \figref{fig:overfit_suppl}(d and f), will have minor discrepancies from the ground truth, as some areas will be smoothed out. 

\subsection{Surface Parametrization}

\begin{figure}[t]
    \centering
    \includegraphics[width=\columnwidth]{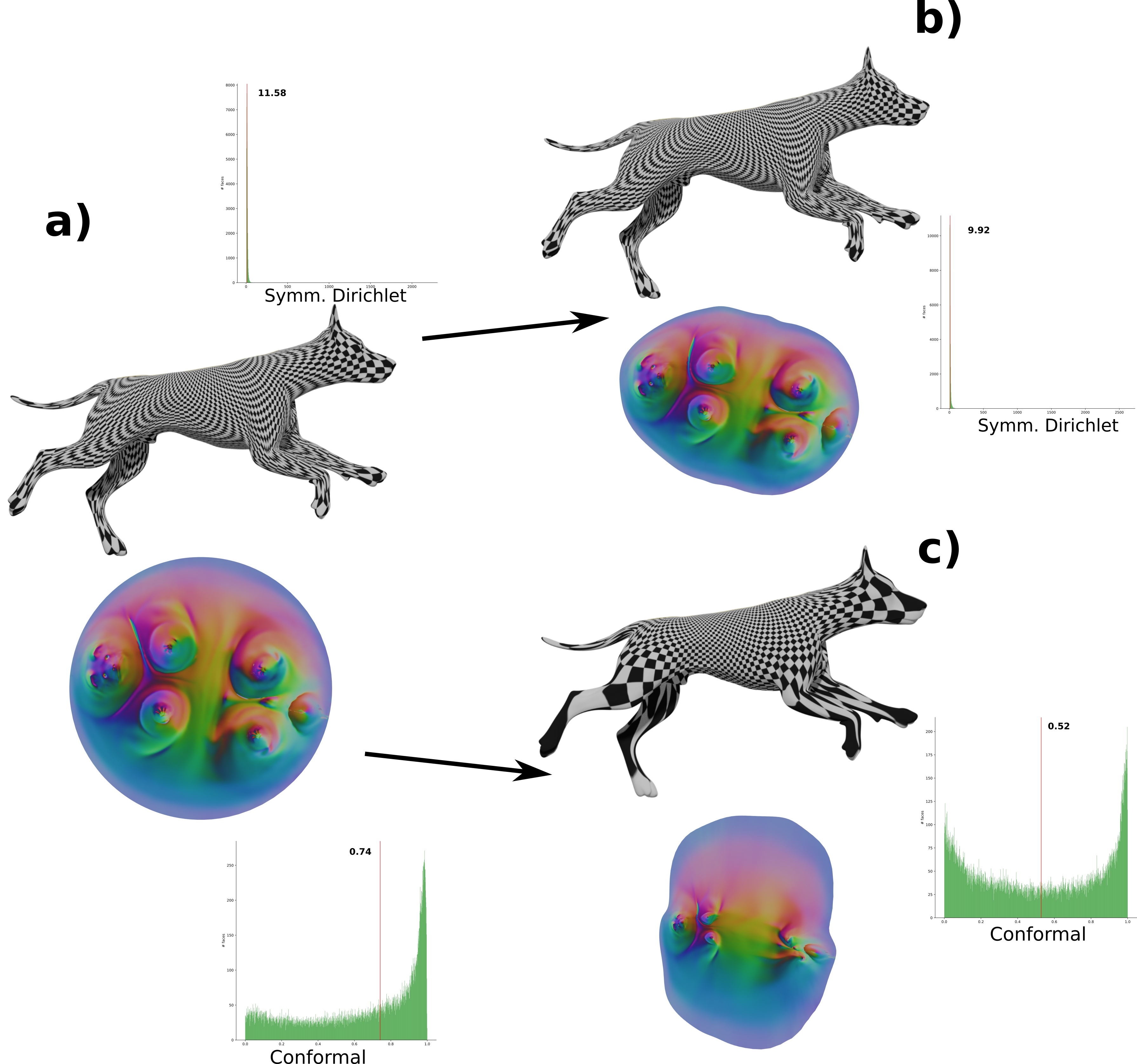}
    \caption{Parametrization with different energies. Starting from a neural surface map (a), we optimize for conformal distortion (b) and isometric distortion boundary free (c).}
    \label{fig:param}
\end{figure}

We evaluate the efficacy of our method in optimizing different energies, by optimizing the distortion of the map $\phi\circ h$ as a function of $h$. We show a parameterization minimizing conformal-distortion in \figref{fig:param}(b), reducing the conformal distortion from $1.74$ to $1.52$. Similarly, we show a  parametrization minimizing isometric distortion in \figref{fig:param}(c) reducing the median symmetric dirichlet from $11.58$ to $9.92$.

\subsection{Composition with Analytical Maps}
\begin{figure}[t]
    \centering
    \includegraphics[width=\columnwidth]{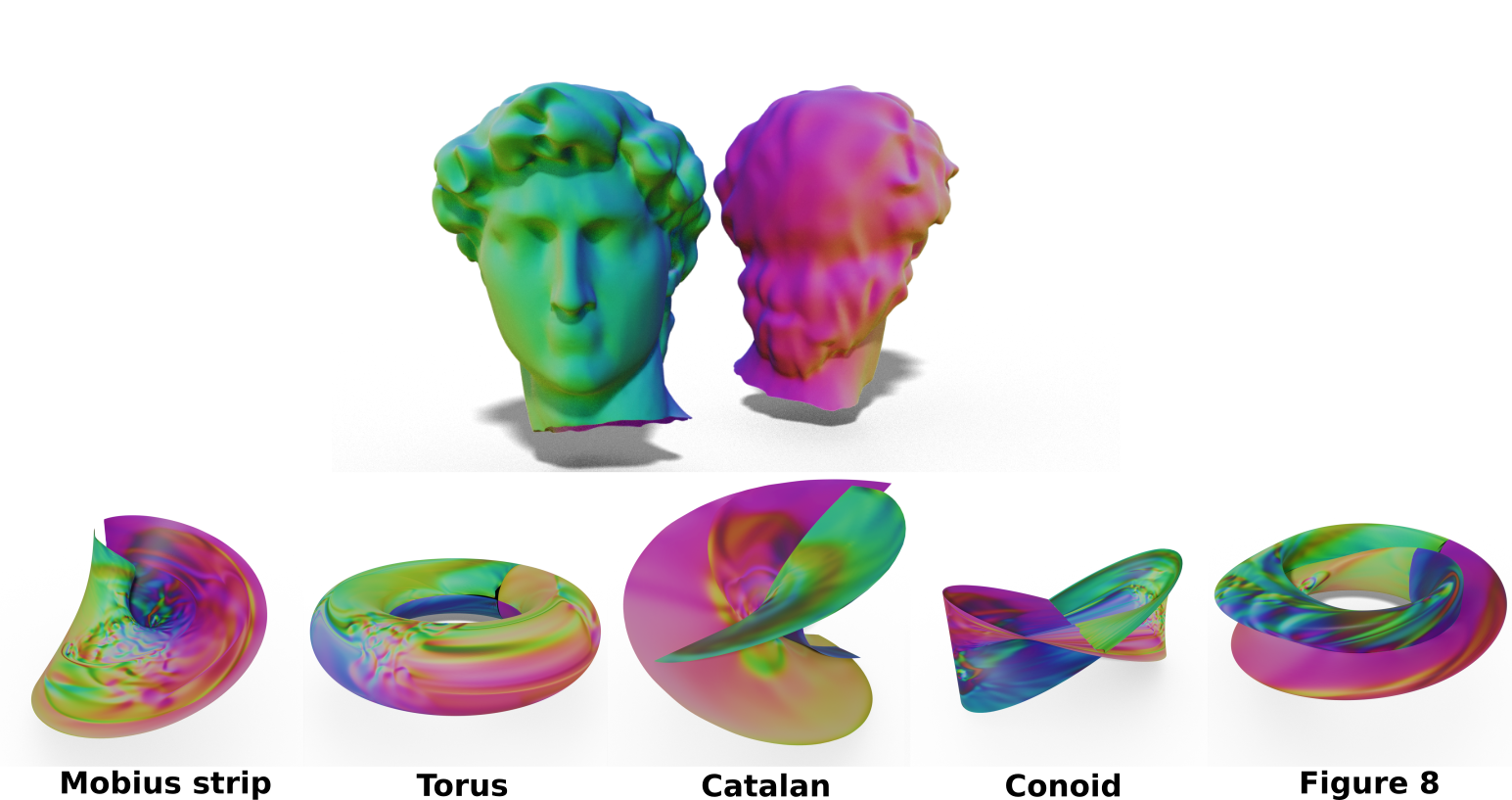}
    \caption{Surface mapping between David and a variety of parametric surfaces. Shape coloring is based on source model normals, \ie, David. }
    \label{fig:analytical}
\end{figure}

As discussed in the paper, we can directly compose a neural map with analytical functions to get surface maps into analytical surfaces. \figref{fig:analytical} illustrates several additional analytic surfaces we can map into. The neural map is optimized to minimize the conformal energy.

\end{document}